\documentclass[runningheads]{llncs}
\usepackage[T1]{fontenc}
\usepackage{graphicx}
\usepackage{booktabs}
\usepackage{multirow}
\usepackage{amsmath,amssymb,amsfonts}
\usepackage{caption}
\usepackage{xcolor}
\usepackage[misc]{ifsym} 
\usepackage[colorlinks=true,citecolor=blue,urlcolor=blue]{hyperref}
\usepackage{bbding}

\begin{document}
\title{FM-OSD: Foundation Model-Enabled One-Shot Detection of Anatomical Landmarks}
\titlerunning{FM-OSD: Foundation Model-Enabled One-Shot Detection of Anatomical}
%
\author{Juzheng Miao\inst{1} \and
Cheng Chen\inst{2}\textsuperscript{(\Letter)} \and
Keli Zhang\inst{3} \and Jie Chuai\inst{3} \and \\Quanzheng Li\inst{2,4} \and Pheng-Ann Heng\inst{1,5}}
%
\authorrunning{J. Miao et al.}
%
\institute{Department of Computer Science and Engineering,\\
The Chinese University of Hong Kong, Hong Kong, China \and
Center for Advanced Medical Computing and Analysis, Massachusetts General\\
Hospital and Harvard Medical School, Boston, MA, USA\\
\email{cchen101@mgh.harvard.edu} \and
Huawei Noah’s Ark Lab, Shenzhen, China \and
Data Science Office, Massachusetts General Brigham, Boston, MA, USA \and
Institute of Medical Intelligence and XR,\\
The Chinese University of Hong Kong, Hong Kong, China
}
\maketitle              
\begin{abstract}
One-shot detection of anatomical landmarks is gaining significant attention for its efficiency in using minimal labeled data to produce promising results.
However, the success of current methods heavily relies on the employment of extensive unlabeled data to pre-train an effective feature extractor, which limits their applicability in scenarios where a substantial amount of unlabeled data is unavailable.
In this paper, we propose the first foundation model-enabled one-shot landmark detection (FM-OSD) framework for accurate landmark detection in medical images by utilizing solely a single template image without any additional unlabeled data.
Specifically, we use the frozen image encoder of visual foundation models as the feature extractor, and introduce dual-branch global and local feature decoders to increase the resolution of extracted features in a coarse-to-fine manner.
The introduced feature decoders are efficiently trained with a distance-aware similarity learning loss to incorporate domain knowledge from the single template image. 
Moreover, a novel bidirectional matching strategy is developed to improve both robustness and accuracy of landmark detection in the case of scattered similarity map obtained by foundation models.
We validate our method on two public anatomical landmark detection datasets. By using solely a single template image, our method demonstrates significant superiority over strong state-of-the-art one-shot landmark detection methods.
Code is available at: \url{https://github.com/JuzhengMiao/FM-OSD}.

\end{abstract}

\section{Introduction}
Accurate anatomical landmark detection is an essential step in many clinical applications, such as disease diagnosis~\cite{jiang2022cephalformer,wang2016benchmark,zhang2017alzheimer}, therapy planning~\cite{bier2018x,yang2015automated}, and registration initialization~\cite{han2014robust,oktay2016stratified,yan2022sam}.
Although deep learning methods have achieved impressively accurate detection results~\cite{jiang2022cephalformer,payer2019integrating,zhou2021learn,zhu2021you}, a large number of high-quality labeled data are typically needed, which are extremely time-consuming and difficult to obtain from domain experts, limiting their use in clinical practice.

Considering these challenges, accurate landmark detection with limited labeled data and even under the one-shot setting, i.e., utilizing one labeled template image, has been explored by researchers~\cite{bai2023samv2,quan2022images,yan2022sam,yao2021one,yin2022one,zhu2023uod}.
These methods typically rely on a good feature extractor to obtain the feature representation for each pixel, based on which a matching strategy, such as the nearest neighbor searching strategy, can be applied to find corresponding positions on the query image given the template image~\cite{bai2023samv2,quan2022images,yan2022sam,yao2021one,zhu2023uod}.
Therefore, current one-shot medical landmark detection methods often leverage network weights pre-trained on natural images and focus on the design of effective contrastive loss to obtain a feature extractor by using large amounts of unlabeled target data.
For example, Yao et al.~\cite{yao2021one} apply the contrastive loss on different network layers by taking the point after augmentation as the positive sample and all others as negative samples.
This method leverages hundreds of unlabeled images in addition to the labeled template image to effectively train the feature extractor.
Zhu at al.~\cite{zhu2023uod} even optimize their feature extractor on a combination of three landmark detection datasets with around 1000 X-ray images for a universal one-shot landmark detection.
Moreover, the unlabeled data are used to train an additional heatmap regression network under the guidance of pseudo labels generated by matching the features from the feature extractor, as discussed in~\cite{quan2022images,yan2022sam,yao2021one,zhu2023uod}.
Therefore, existing one-shot anatomical landmark detection methods are limited by their dependence on large quantities of unlabeled data, which constrains their practical use in scenarios where additional images are unavailable.

To eliminate the reliance on a substantial volume of unlabeled data for one-shot anatomical landmark detection, we are motivated to investigate the efficacy of pre-trained foundation models as feature extractors. 
This inspiration is rooted in the remarkable zero-shot capabilities of foundation models like DINO~\cite{caron2021emerging}, DINOv2~\cite{oquab2023dinov2}, and SAM~\cite{kirillov2023segment}, demonstrated in various tasks~\cite{amir2021deep,anand2023one}.
However, leveraging features extracted from foundation models for landmark detection in medical images faces multiple challenges. 
\textbf{First}, the resolution of features from foundation models is inherently lower because the transformer architecture provides patch-wise encoding rather than pixel-wise representation. Yet, for landmark detection tasks, high-resolution features are crucial~\cite{amir2021deep,an2023sharpose}. 
\textbf{Second}, large foundation models are usually pre-trained on natural images, which can impact their ability to discriminate effectively when extracting features from medical images due to the domain gap~\cite{deng2023segment,he2023accuracy}.
\textbf{Third}, precise landmark detection requires accurate position-related semantic information, which is often ambiguous in features from foundation models like DINO~\cite{amir2021deep}. This ambiguity is exacerbated in medical images, where similar patches are widespread, potentially leading to a scattered similarity map and causing the commonly used matching strategy such as nearest neighbor search method to identify incorrect points.


In this paper, to the best of our knowledge, we present the first \textbf{f}oundation \textbf{m}odel-enabled \textbf{o}ne-\textbf{s}hot landmark \textbf{d}etection (FM-OSD) framework for medical images, achieving accurate landmark detection by utilizing only a single template image, without requiring any additional unlabeled images.  
Our method roots in the powerful feature extraction capabilities of pre-trained visual foundation models, utilizing their frozen image encoder as our feature extractor.
Considering the challenges in the features directly extracted from foundation models on medical images, we first introduce dual-branch global and local feature decoders to enhance the resolution of extracted features for landmark detection in a coarse-to-fine manner. 
The introduced feature decoders are lightweight and learnable, and can be efficiently trained with a proposed distance-aware similarity learning loss to integrate domain knowledge contained in the one labeled template image. 
Moreover, we develop a bidirectional matching strategy to improve both robustness and accuracy of landmark detection in the case of scattered similarity maps, by taking the inverse matching error on the template image as a guidance.
Our method demonstrates significant improvements over state-of-the-art (SOTA) methods for one-shot landmark detection on two public X-ray datasets.
Our FM-OSD method obtains an improvement of over 16\% in terms of the detection error, compared with various strong baselines that require hundreds of unlabeled data from the target dataset.

\section{Method}
Fig. 1 illustrates our one-shot anatomical landmark detection framework with foundation models. By leveraging the powerful feature extraction capabilities of visual foundation models and our  proposed novel feature enhancement and matching strategies, our method achieves anatomical landmark detection using only a single template image, without requiring any additional unlabeled images. 
\subsection{Global and Local Feature Enhancement of Foundation Model}
We aim to leverage deep features extracted from a pre-trained visual foundation model as dense visual descriptors for one-shot landmark detection tasks. 
However, the image encoder of a pre-trained foundation model often generate feature maps with downsampled resolution, which significantly restricts their effectiveness in landmark detection. 
Additionally, the feature extractor of the foundation model is trained on natural images, leading to a substantial domain shift when applied to medical images. 
To address these challenges, we propose a coarse-to-fine landmark detection framework featuring dual-branch global and local learnable decoders.
Our approach enhances the feature resolution for landmark detection in a coarse-to-fine manner and improves feature quality by incorporating domain knowledge contained in the template image.


\begin{figure}[!t]
 \centering
 \includegraphics[width=0.95\textwidth]{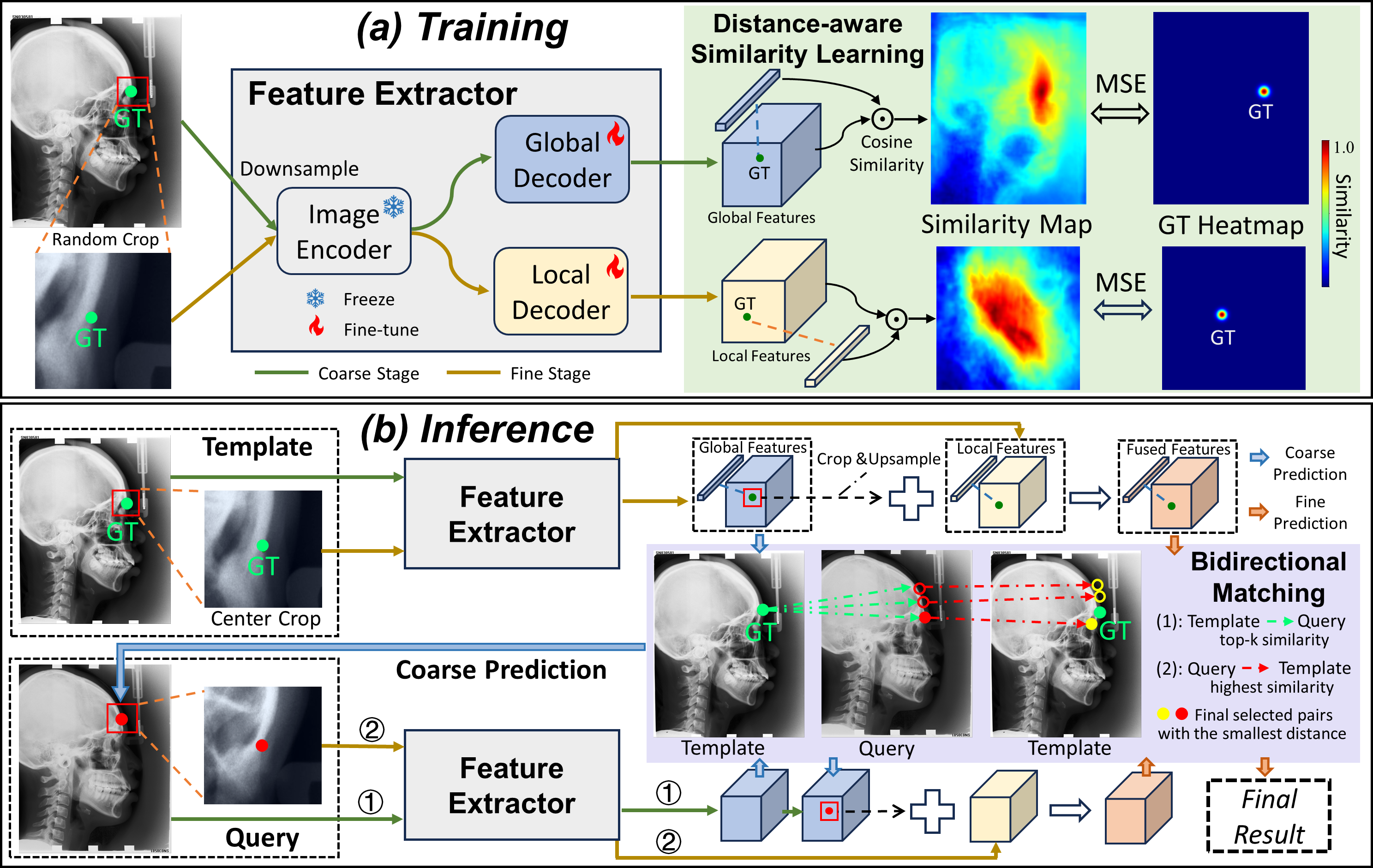}
 \caption{Overview of our proposed method. In training, two light decoders are updated using the distance-aware similarity learning, while the features are frozen and a bidirectional matching strategy on top of the combination of global and local features is adopted to find robust landmark predictions on query images.} \label{method}
\end{figure}
Specifically, as shown in Fig. 1(a), for the coarse stage, the entire input image $I\in {\mathcal{R}^{H \times W}}$ is downsampled into a size with the short side of 224 and is then fed into the frozen image encoder $\mathcal{E}$ of a foundation model, which typically undergoes additional downsampling of the feature maps due to the patch encoding process in vision transformers. In order to restore the feature resolution that is diminished during the patch encoding process, a trainable global feature decoder $\mathcal{D}_G$ consisting of upsampling and convolutional layers is attached to the image encoder, facilitating the extraction of global features, denoted by $F^G$. 
The coarse landmark positions on the query images can then be identified by matching the similarity of global features between the query images and the template image with our proposed bidirectional matching strategy for which we will elaborate in the next section: $P_c = BDM(F^{G,t},F^{G,q})$, where $F^{G,t}$ and $F^{G,q}$ represent the global features of the template and query image, respectively.
For the fine stage, we aim to construct high-resolution local representations by cropping a 224$\times$224 local region $I_{i}^L$ containing the identified coarse position for each landmark, with $i$ incidating the $i$-th landmark. The cropped local region is fed into the frozen image encoder and a local feature decoder $\mathcal{D}_L$ consisting of multiple deconvolutional layers to obtain fine-grained local features $F_i^L$. 
To further incorporate both the global context information into the local features, we upsample the corresponding global feature regions to the same size of the local feature maps and add them together to obtain the fused local features $F_i^F$ for final matching.


\textbf{Distance-aware similarity learning loss.} To enhance the feature quality extracted from the foundation model with increased discrimination among features from different positions, we aim to effectively optimize the global and local feature decoders by leveraging just one single template image $I_t\in {\mathcal{R}^{H \times W}}$ annotated with $N$ landmarks $P_t=\{(x_i^t, y_i^t) \mid i=1, \cdots, N\}$.
Intuitively, the feature of points close to the target landmark should have a high similarity to the feature of the target landmark, and the points far away from the target landmark should have a low similarity.
Therefore, we takes the distance to the target landmark into consideration and proposes a distance-aware similarity learning loss. 
The prediction similarity map is obtained by calculating the cosine similarity between the feature of the ground truth landmark and the feature of other points across the image: $S_i=\cos(F,F\left(x_i^t, y_i^t\right))$, where $F$ indicates the global or local feature map and $F\left(x_i^t, y_i^t\right)$ means the feature vector on $F$ at the position of $\left(x_i^t, y_i^t\right)$. "cos" denotes the cosine similarity.
The ground truth similarity map for this landmark is generated by using a 2D Gaussian distribution as follows:
$Y_i=e^{-\frac{\left(x-x_i^t\right)^2+\left(y-y_i^t\right)^2}{2 \sigma^2}}, x \in [0,H-1], y \in [0,W-1]$
,where the center of the Gaussian distribution is located at the ground truth landmark $(x_i^t,y_i^t)$, with the highest similarity of 1.
The ground truth similarity value decreases with the increase of the distance to the ground truth landmark, where the decreasing speed is controlled by the standard deviation $\sigma$.
Finally, a Mean Square Error (MSE) constraint is imposed between the global/local similarity maps ($S_{i}^G,S_i^L$) and their corresponding ground truth heatmaps ($Y_{i}^G,Y_i^L$) for all the $N$ landmarks:
\begin{equation}
\small
L=\frac{1}{N} \sum_{i=1}^N MSE\left(S_{i}^G, Y_{i}^G\right)+MSE\left(S_i^L, Y_i^L\right)
\end{equation}


\subsection{Bidirectional Matching Strategy}
With the extracted features of the template and query image, the success of one-shot landmark detection relies on the development of an effective matching strategy to find the corresponding point on the query image based on feature similarity with the ground truth landmark on the template image.
The commonly used matching strategy is selecting the point with highest feature similarity to the template landmark.
However, the feature similarity map often lacks the accurate focus on a single point since the feature extracted by foundation models like DINO is usually ambiguous~\cite{amir2021deep} in terms of their spatial distribution, reflecting similar features over extensive areas, as shown in the similarity map in Fig.~\ref{method}.
As a result, directly choosing the query point with the highest similarity to template landmark could result in inaccurate landmark detection.

To improve the matching accuracy, we propose a novel bidirectional matching strategy (BDM) to find the landmark pair $(\hat{ \boldsymbol{p}}_i^q,\hat{\boldsymbol{p}}_i^t)$ with not only a high feature similarity from template to query image but also a low inverse-matching error from query to template. $(\hat{ \boldsymbol{p}}_i^q,\hat{\boldsymbol{p}}_i^t)$ indicates the matched position for the $i$-th landmark on the query image and the template image, respectively and $\hat{ \boldsymbol{p}}_i^q$ is the final prediction for the query image.
Such a two-side agreement is similar to the Best-Buddies Similarity method for template matching~\cite{dekel2015best}, which has been theoretically and experimentally demonstrated to be a robust and reliable matching method between two sets of points.
Moreover, the inverse-matching error for the matching from query to template can be calculated since the ground truths are known and this can be taken as an estimation of the matching error on the query image $I_q$.
Specifically, our proposed BDM is formulated as follows:
\begin{equation}
\small
\begin{gathered}
(\hat{ \boldsymbol{p}}_i^q,\hat{\boldsymbol{p}}_i^t)=\underset{(\boldsymbol{c}_i^q, \boldsymbol{c}_i^t) \in \Omega}{\operatorname{argmin}} \operatorname{Dist}(\boldsymbol{c}_i^t, \boldsymbol{p}_i^t), i=1, \cdots, N \\
\text { s.t. } \Psi=\{\boldsymbol{c}_i^q \mid \boldsymbol{c}_i^q\in\underset{(x, y) \in I_q}{\operatorname{argtopk}} \cos(F^q, F^t(\boldsymbol{p}_i^t))\} \\
\Omega=\{(\boldsymbol{c}_i^q, \boldsymbol{c}_i^t) \mid \boldsymbol{c}_i^q \in \Psi, \boldsymbol{c}_i^t=\underset{(x, y) \in I_t}{\operatorname{argmax}} \cos(F^t, F^q(\boldsymbol{c}_i^q))\}
\end{gathered}
\end{equation}
where $\boldsymbol{p}_i^t=\left(x_i^t, y_i^t\right) \in P_t$ is the annotation on $I_t$ and "Dist" calculates the Euclidean distance.
"argtopk" selects the points with the top-$k$ similarity values in the query image $I_q$.
$F^q$ and $F^t$ are the feature map of $I_q$ and $I_t$ and are set as $F^{G,q},F^{G,t}$ and $F_i^{F,q},F_i^{F,t}$ for the coarse and fine stage, respectively.
$F^t(\boldsymbol{p}_i^t),F^q(\boldsymbol{c}_i^q)$ denote the feature vector on $F^t,F^q$ with a position of $\boldsymbol{p}_i^t, \boldsymbol{c}_i^q$.

There is a total of three steps to solve this optimization problem.
First, find a set of $k$ candidate points on the query image $I_q$ with the top-$k$ highest similarities to the template landmark feature $F^t(\boldsymbol{p}_i^t)$, denoted as $\Psi$.
Second, for each point in $\Psi$, we find its matching point $\boldsymbol{c}_i^t$ back to the template image $I_t$ with the highest similarity and include each candidate point pair in $\Omega$.
Finally, we calculate the distance between $\boldsymbol{c}_i^t$ and the ground truth landmark $\boldsymbol{p}_i^t$ for each pair in $\Omega$ and select the one with the smallest distance, i.e., $(\hat{ \boldsymbol{p}}_i^q,\hat{\boldsymbol{p}}_i^t)$. The corresponding point on the query image $\hat{ \boldsymbol{p}}_i^q$ is taken as the final predicted landmark.
The BDM method is applied to the matching of both global features and local features.

\section{Experimental Results}
\textbf{Datasets and Evaluation Metrics.}
We evaluate the effectiveness of our method using two public X-ray datasets that are commonly utilized in prior one-shot anatomical landmark detection studies.
The Head X-ray dataset~\cite{wang2016benchmark} consists of 400 lateral cephalograms with a resolution of 0.1 mm and 19 target landmarks.
This dataset has been officially split, and we use a single labeled image from the training set as our template image, while employing the 250 testing images as our test set. 
The Hand X-ray dataset, labeled by~\cite{payer2019integrating}, contains 909 radiographs with 37 labeled landmarks. The length between two endpoints of the wrist is assumed to be 50 mm~\cite{payer2019integrating}.
Following~\cite{payer2019integrating,quan2022images,yin2022one}, we use one image from the first 609 images as the template image, and utilize the remaining images for testing.
Two commonly used evaluation metrics are employed, i.e., the mean radial error (MRE) and the successful detection rate (SDR)~\cite{quan2022images,yin2022one}.
The MRE measures the Euclidean distance between two points while SDR counts the percentage of predictions with a distance under various thresholds, which are set to be 2 mm, 2.5 mm, 3 mm, 4mm for the Head X-ray dataset and 2 mm, 4 mm, 10 mm for the Hand X-ray dataset, respectively.

\noindent \textbf{Implementation Details.}
The image encoder in our method is taken from DINO-S~\cite{caron2021emerging}, with features extracted from the "key" head of the ninth layer.
The investigation on using different foundation models and different layers have been provided in our ablation study. 
The template image is augmented with random shifting, scaling and rotating for 500 times.
Following~\cite{amir2021deep}, the patch size and the stride are set as 8$\times$8 and 4, respectively for patch generation.
The two decoders have an output feature dimension of 256 and are optimized by an Adam optimizer with a fixed learning rate of 2e-4 and a batch size of 4.
On the Head dataset, the global decoder is updated by 20000 iterations using a standard deviation of 5 for the ground-truth Gaussian similarity map, while the local decoder is updated by 1000 iterations with a standard deviation of 2.
On the Hand dataset, we optimize both decoders using a standard deviation of 8 but update the global and local decoder by 20000 and 3000 iterations, respectively.
During inference, the $k$ for BDM is empirically set as 3 and 5 for the Head and Hand dataset, respectively.
Our method is implemented using Pytorch and trained on an NVIDIA A40 GPU.

\noindent \textbf{Comparisons with State-of-the-arts.}
We compare our method with SOTA one-shot anatomical landmark detection methods, including CC2D~\cite{yao2021one}, SAEM \cite{yan2022sam}, SCP~\cite{quan2022images}, EGTNLR~\cite{yin2022one} and UOD~\cite{zhu2023uod}.
All these comparison approaches and our method use the same template image as in~\cite{yao2021one,yao2022relative,zhu2023uod}, with the difference being that our method does not employ additional unlabeled images. 
We also compare with SOTA fully supervised methods GU2Net~\cite{zhu2021you} as an upper bound.
The results for other methods are directly obtained from their original papers since we use the same datasets and template image, except for SAEM and EGTNLR, for which we implement their released code on the two datasets.

\begin{table}[!t]
\caption{Comparisons with SOTA methods on the Head and Hand dataset. "Label" and "Unlabel" denote the number of labeled and unlabeled data, respectively. $^+$ indicates the method is trained with multiple datasets.}
\label{compare}
\centering
\scriptsize
\resizebox{0.95\linewidth}{!}{
\begin{tabular}{c|c|c|c|c|c|c||c|c|c|c|c}
\toprule[1.0pt]
\multirow{3}{*}{Method}    & \multicolumn{6}{c||}{\textbf{Head Dataset}}                                                                                                                                                                                                  & \multicolumn{5}{c}{\textbf{Hand Dataset}}                                                                                                                                                                     \\ \cline{2-12} 
                           & {\multirow{2}{*}{\begin{tabular}[c]{@{}c@{}}Label/\\ Unlabel\end{tabular}}} & {MRE↓} & \multicolumn{4}{c||}{SDR (\%)↑}                                                                & {\multirow{2}{*}{\begin{tabular}[c]{@{}c@{}}Label/\\ Unlabel\end{tabular}}} & {MRE↓} & \multicolumn{3}{c}{SDR (\%)↑}                                   \\ \cline{4-7} \cline{10-12} 
                           & {}                                                                          & {(mm)} & {2 mm}  & {2.5 mm} & {3 mm}  & 4 mm  & {}                                                                          & {(mm)} & {2 mm}  & {4 mm}  & 10 mm \\ \hline
                           \hline
GU2Net~\cite{zhu2021you}         & {988/0$^+$}                                                                   & {1.54} & {77.79} & {84.65}  & {89.41} & 94.93 & {988/0$^+$}                                                                   & {0.84} & {95.40}  & {99.35} & 99.75 \\ \hline
CC2D~\cite{yao2021one}          & {1/149}                                                                     & {2.36} & {51.81} & {63.13}  & {73.66} & 86.25 & {1/608}                                                                     & {2.65} & {51.19} & {82.56} & 95.62 \\
SAEM~\cite{yan2022sam}             & {1/149}                                                                     & {2.58} & {54.34} & {64.51}  & {70.82} & 80.76 & {1/608}                                                                     & {1.69} & {76.61} & {92.52} & 99.08 \\
SCP~\cite{quan2022images}           & {1/149}                                                                     & {2.74} & {43.79} & {53.05}  & {64.12} & 79.05 & {1/608}                                                                     & {2.47} & {-} & {-} & - \\
EGTNLR~\cite{yin2022one} & {1/149}                                                                     & {2.27} & {49.45} & {63.07}  & {74.70}  & 88.91 & {1/608}                                                                     & {1.81} & {64.62} & {95.03} & \textbf{99.97} \\
UOD~\cite{zhu2023uod}         & {3/985$^+$}                                                                    & {2.43} & {51.14} & {62.37}  & {74.40}  & 86.49 & {3/985$^+$}                                                                    & {2.52} & {53.37} & {84.27} & 97.59 \\
\hline
FM-OSD (ours)                       & {1/0}                                                                       & {\textbf{1.82}} & {\textbf{67.35}} & {\textbf{77.92}}  & {\textbf{84.59}} & \textbf{91.92} & {1/0}                                                                       & {\textbf{1.41}} & {\textbf{86.66}}  & {\textbf{96.66}} & 99.11 \\ 
\toprule[1.0pt]
\end{tabular}
}
\end{table}

\begin{figure}[!t]
 \centering
 \includegraphics[width=0.9\textwidth]{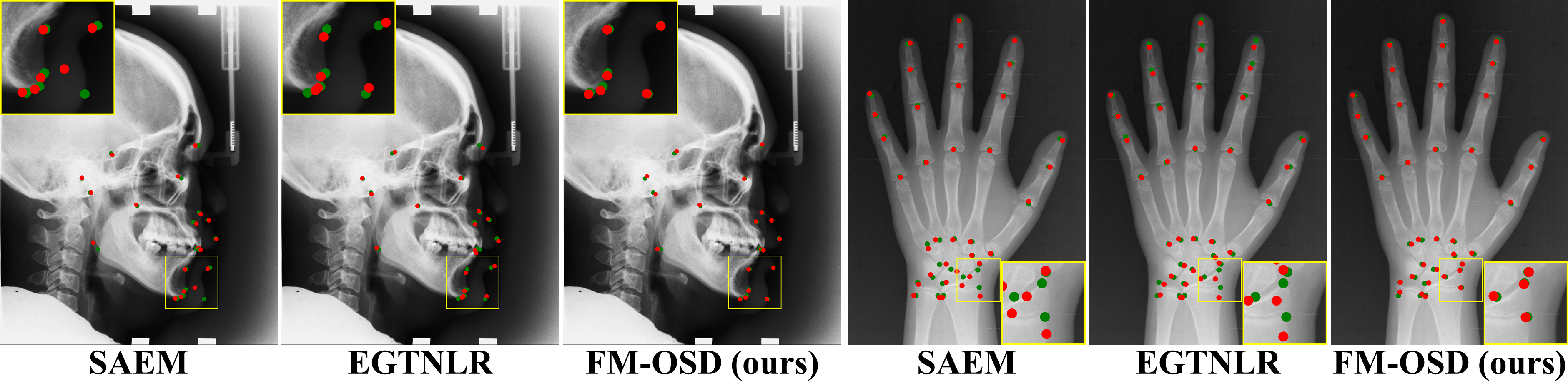}
 \caption{Visualizations of different methods on the head and hand dataset. Red and green points indicate predicted landmarks and ground-truth labels, respectively.} \label{visual}
\end{figure}
As shown in Table~\ref{compare}, our FM-OSD method significantly outperforms other SOTA one-shot methods on both datasets, even though we only utilize one template image without leveraging additional unlabeled data as those methods do~\cite{quan2022images,yan2022sam,yin2022one}.
On the Head dataset, our method obtains an MRE of 1.82 mm, which is over 19\% superior to EGTNLR method with an MRE of 2.27 mm.
Also, our method outperforms other one-shot methods by a margin of more than 10\% in terms of the SDR under 2 mm, 2.5 mm, and 3 mm.
Similar results can be found on the Hand dataset, where we achieve an improvement of at least 16\% in terms of MRE (1.41 mm vs 1.69 mm) and more than 10\% for the SDR under 2 mm. The much higher SDR results under small thresholds indicate that our method can localize more landmarks in a low error.
Fig.~\ref{visual} shows that the predictions from our FM-OSD method are closer to the ground truth points.


\begin{figure}[!tbp]
\begin{minipage}[b]{.65\linewidth}
\centering
\scriptsize
\begin{tabular}{c|c|c|c|c|c|c}
\hline
\multirow{2}{*}{BDM} & \multirow{2}{*}{Global} & \multirow{2}{*}{Local} & \multirow{2}{*}{MRE (mm)↓} & \multicolumn{3}{c}{SDR (\%)↑}  \\
\cline{5-7}
& & & & 2 mm  & 4 mm & 10 mm  \\
\hline
& & & 3.72 & 31.13 & 77.49  & 97.60 \\
\checkmark & & & 2.82 & 51.87 & 89.06  & 98.23 \\
\checkmark  & \checkmark & & 1.86 & 76.75 & 96.60 & \textbf{99.13} \\
\checkmark & \checkmark & \checkmark & \textbf{1.41} & \textbf{86.66} & \textbf{96.66}  & 99.11\\
\hline
\end{tabular}
\captionof{table}{Ablation studies of different components of our method on the hand dataset.}
\label{tab:ablation}
\end{minipage}
\begin{minipage}[b]{.3\linewidth}
\centering
\includegraphics[width=\textwidth, height=0.5\textwidth]{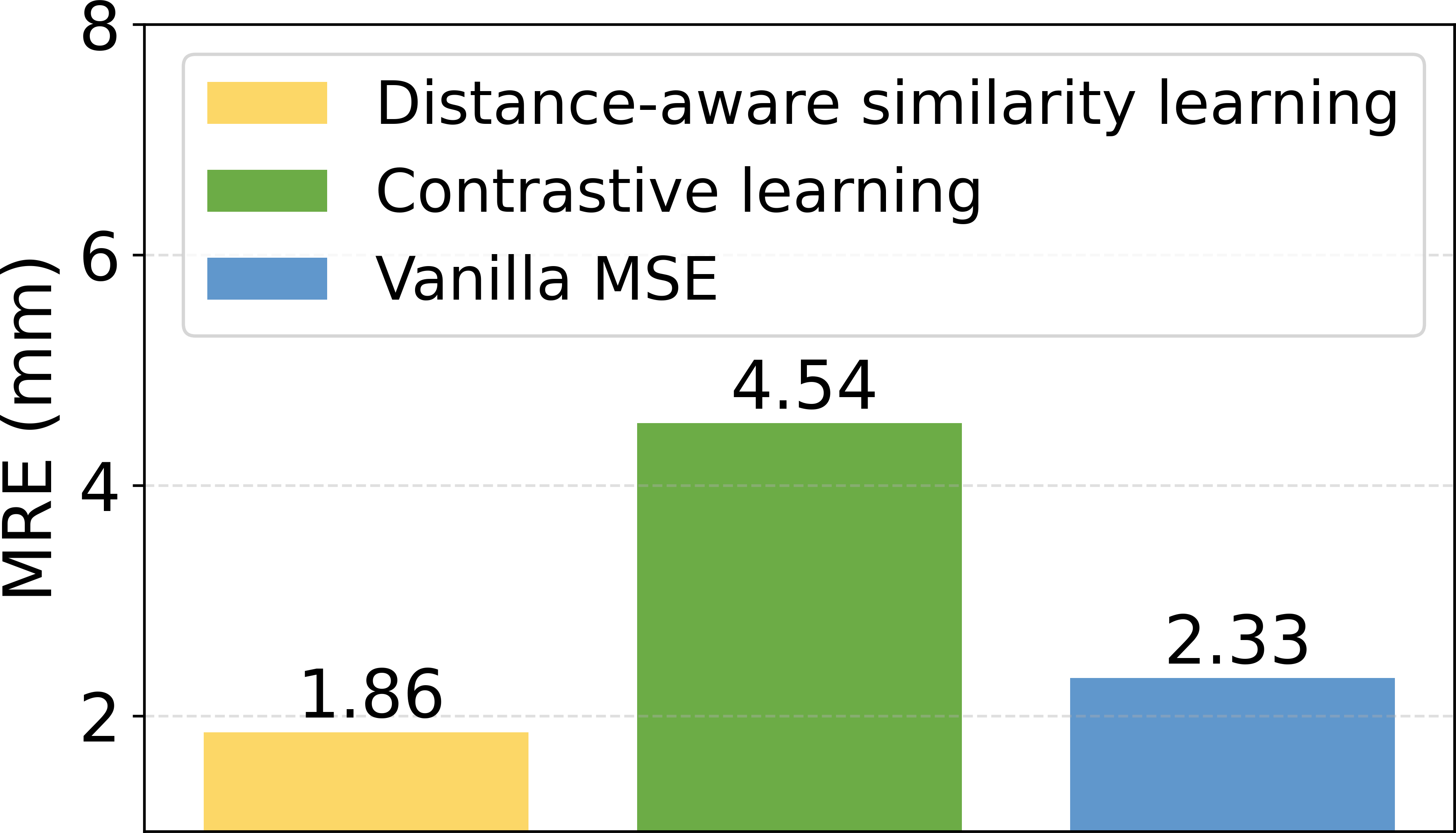}
\caption{Effects of various losses for training $\mathcal{D}_G$.}
\label{fig:ablation}
\end{minipage}
\end{figure}

\begin{figure}[!t]
 \centering
 \includegraphics[width=0.95\textwidth]{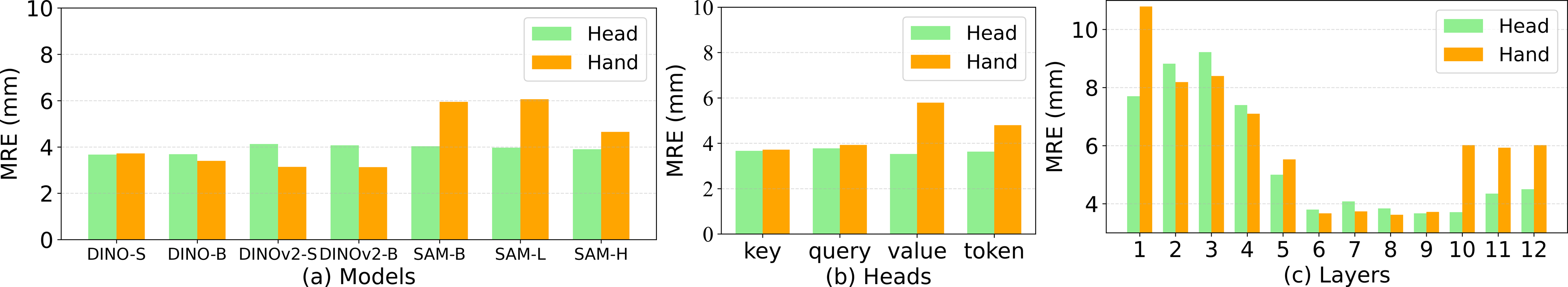}
 \caption{Zero-shot performances using various (a) models, (b) heads, and (c) layers.} \label{abla_studies}
\end{figure}

\noindent \textbf{Ablation Studies.}
Table~\ref{tab:ablation} shows ablation studies on the key components of our method with the Hand dataset.
Since BDM is a inference strategy for feature similarity matching without the need of training, we first apply it to the original features extracted by the frozen image encoder of foundation models.
Leveraging our proposed BDM strtegy, the performance significantly improves by 0.9 mm in MRE.
By introducing the global decoder and the local decoder under a coarse-to-fine scheme, the MRE can be reduced to 1.86 mm and 1.41 mm, respectively.
Moreover, to show the efficacy of the distance-aware similarity learning loss, we replace it with a contrastive learning loss by taking all the pixels on the feature map except the target point itself as negative samples and an MSE loss posed on the similarity map guided by a ground truth map whose value is 1 for the target point and 0 for others (See Fig.~\ref{fig:ablation}).
The MRE results for the training of the global decoder on the hand dataset are 4.54 mm and 2.33 mm for the two losses, respectively, whereas using our proposed loss can obtain an MRE of 1.86 mm.
Fig.~\ref{abla_studies} (a) compares the results of various image encoders from several popular foundation models.
DINO-S is chosen in this work considering its stable performance on both datasets and the small model size, but our proposed method can be applied to others without modifications.
We also compare the impacts of which transformer head and which layer the feature is extracted from in Fig.~\ref{abla_studies} (b) and (c). 
The transformer heads include the key, query, value and token (the output of a transformer block).
Generally, the key head and the ninth layer can obtain accurate results on both datasets, and are thus used in our work.

\section{Conclusion}This paper proposes a foundation model-enabled one-shot landmark detection framework for medical images.
With careful designs to tackle the challenges of adapting foundation models to landmark detection, our proposed method can obtain significantly superior results using solely a single template image than other one-shot methods which even use plenty of unlabeled data.
However, we only validate the efficacy of our method on 2D images.
More evaluations on 3D datasets and cross-modality settings will be conducted in our future studies.

\begin{credits}
\subsubsection{\ackname} The work described in this paper was supported in part by the Research Grants Council of the Hong Kong Special Administrative Region, China, under Project T45-401/22-N; and by the Hong Kong Innovation and Technology Fund (Project No. MHP/085/21).

\end{credits}

\bibliographystyle{splncs04}
\bibliography{Paper-0320}

\begin{thebibliography}{10}
\providecommand{\url}[1]{\texttt{#1}}
\providecommand{\urlprefix}{URL }
\providecommand{\doi}[1]{https://doi.org/#1}

\bibitem{amir2021deep}
Amir, S., Gandelsman, Y., Bagon, S., Dekel, T.: Deep vit features as dense visual descriptors. arXiv preprint arXiv:2112.05814  \textbf{2}(3), ~4 (2021)

\bibitem{an2023sharpose}
An, X., Zhao, L., Gong, C., Wang, N., Wang, D., Yang, J.: Sharpose: Sparse high-resolution representation for human pose estimation. arXiv preprint arXiv:2312.10758  (2023)

\bibitem{anand2023one}
Anand, D., Singhal, V., Shanbhag, D.D., KS, S., Patil, U., Bhushan, C., Manickam, K., Gui, D., Mullick, R., Gopal, A., et~al.: One-shot localization and segmentation of medical images with foundation models. arXiv preprint arXiv:2310.18642  (2023)

\bibitem{bai2023samv2}
Bai, X., Bai, F., Huo, X., Ge, J., Lu, J., Ye, X., Yan, K., Xia, Y.: Samv2: A unified framework for learning appearance, semantic and cross-modality anatomical embeddings. arXiv preprint arXiv:2311.15111  (2023)

\bibitem{bier2018x}
Bier, B., Unberath, M., Zaech, J.N., Fotouhi, J., Armand, M., Osgood, G., Navab, N., Maier, A.: X-ray-transform invariant anatomical landmark detection for pelvic trauma surgery. In: International Conference on Medical Image Computing and Computer-Assisted Intervention. pp. 55--63. Springer (2018)

\bibitem{caron2021emerging}
Caron, M., Touvron, H., Misra, I., J{\'e}gou, H., Mairal, J., Bojanowski, P., Joulin, A.: Emerging properties in self-supervised vision transformers. In: Proceedings of the IEEE/CVF international conference on computer vision. pp. 9650--9660 (2021)

\bibitem{dekel2015best}
Dekel, T., Oron, S., Rubinstein, M., Avidan, S., Freeman, W.T.: Best-buddies similarity for robust template matching. In: Proceedings of the IEEE conference on computer vision and pattern recognition. pp. 2021--2029 (2015)

\bibitem{deng2023segment}
Deng, R., Cui, C., Liu, Q., Yao, T., Remedios, L.W., Bao, S., Landman, B.A., Wheless, L.E., Coburn, L.A., Wilson, K.T., et~al.: Segment anything model (sam) for digital pathology: Assess zero-shot segmentation on whole slide imaging. arXiv preprint arXiv:2304.04155  (2023)

\bibitem{han2014robust}
Han, D., Gao, Y., Wu, G., Yap, P.T., Shen, D.: Robust anatomical landmark detection for mr brain image registration. In: Medical Image Computing and Computer-Assisted Intervention--MICCAI 2014: 17th International Conference, Boston, MA, USA, September 14-18, 2014, Proceedings, Part I 17. pp. 186--193. Springer (2014)

\bibitem{he2023accuracy}
He, S., Bao, R., Li, J., Grant, P.E., Ou, Y.: Accuracy of segment-anything model (sam) in medical image segmentation tasks. arXiv preprint arXiv:2304.09324  (2023)

\bibitem{jiang2022cephalformer}
Jiang, Y., Li, Y., Wang, X., Tao, Y., Lin, J., Lin, H.: Cephalformer: Incorporating global structure constraint into visual features for general cephalometric landmark detection. In: International Conference on Medical Image Computing and Computer-Assisted Intervention. pp. 227--237. Springer (2022)

\bibitem{kirillov2023segment}
Kirillov, A., Mintun, E., Ravi, N., Mao, H., Rolland, C., Gustafson, L., Xiao, T., Whitehead, S., Berg, A.C., Lo, W.Y., et~al.: Segment anything. arXiv preprint arXiv:2304.02643  (2023)

\bibitem{oktay2016stratified}
Oktay, O., Bai, W., Guerrero, R., Rajchl, M., De~Marvao, A., O’Regan, D.P., Cook, S.A., Heinrich, M.P., Glocker, B., Rueckert, D.: Stratified decision forests for accurate anatomical landmark localization in cardiac images. IEEE transactions on medical imaging  \textbf{36}(1),  332--342 (2016)

\bibitem{oquab2023dinov2}
Oquab, M., Darcet, T., Moutakanni, T., Vo, H., Szafraniec, M., Khalidov, V., Fernandez, P., Haziza, D., Massa, F., El-Nouby, A., et~al.: Dinov2: Learning robust visual features without supervision. arXiv preprint arXiv:2304.07193  (2023)

\bibitem{payer2019integrating}
Payer, C., {\v{S}}tern, D., Bischof, H., Urschler, M.: Integrating spatial configuration into heatmap regression based cnns for landmark localization. Medical image analysis  \textbf{54},  207--219 (2019)

\bibitem{quan2022images}
Quan, Q., Yao, Q., Li, J., Zhou, S.K.: Which images to label for few-shot medical landmark detection? In: Proceedings of the IEEE/CVF Conference on Computer Vision and Pattern Recognition. pp. 20606--20616 (2022)

\bibitem{wang2016benchmark}
Wang, C.W., Huang, C.T., Lee, J.H., Li, C.H., Chang, S.W., Siao, M.J., Lai, T.M., Ibragimov, B., Vrtovec, T., Ronneberger, O., et~al.: A benchmark for comparison of dental radiography analysis algorithms. Medical image analysis  \textbf{31},  63--76 (2016)

\bibitem{yan2022sam}
Yan, K., Cai, J., Jin, D., Miao, S., Guo, D., Harrison, A.P., Tang, Y., Xiao, J., Lu, J., Lu, L.: Sam: Self-supervised learning of pixel-wise anatomical embeddings in radiological images. IEEE Transactions on Medical Imaging  \textbf{41}(10),  2658--2669 (2022)

\bibitem{yang2015automated}
Yang, D., Zhang, S., Yan, Z., Tan, C., Li, K., Metaxas, D.: Automated anatomical landmark detection ondistal femur surface using convolutional neural network. In: 2015 IEEE 12th international symposium on biomedical imaging (ISBI). pp. 17--21. IEEE (2015)

\bibitem{yao2021one}
Yao, Q., Quan, Q., Xiao, L., Kevin~Zhou, S.: One-shot medical landmark detection. In: Medical Image Computing and Computer Assisted Intervention--MICCAI 2021: 24th International Conference, Strasbourg, France, September 27--October 1, 2021, Proceedings, Part II 24. pp. 177--188. Springer (2021)

\bibitem{yao2022relative}
Yao, Q., Wang, J., Sun, Y., Quan, Q., Zhu, H., Zhou, S.K.: Relative distance matters for one-shot landmark detection. arXiv preprint arXiv:2203.01687  (2022)

\bibitem{yin2022one}
Yin, Z., Gong, P., Wang, C., Yu, Y., Wang, Y.: One-shot medical landmark localization by edge-guided transform and noisy landmark refinement. In: European Conference on Computer Vision. pp. 473--489. Springer (2022)

\bibitem{zhang2017alzheimer}
Zhang, J., Liu, M., An, L., Gao, Y., Shen, D.: Alzheimer's disease diagnosis using landmark-based features from longitudinal structural mr images. IEEE journal of biomedical and health informatics  \textbf{21}(6),  1607--1616 (2017)

\bibitem{zhou2021learn}
Zhou, G.Q., Miao, J., Yang, X., Li, R., Huo, E.Z., Shi, W., Huang, Y., Qian, J., Chen, C., Ni, D.: Learn fine-grained adaptive loss for multiple anatomical landmark detection in medical images. IEEE Journal of Biomedical and Health Informatics  \textbf{25}(10),  3854--3864 (2021)

\bibitem{zhu2023uod}
Zhu, H., Quan, Q., Yao, Q., Liu, Z., Zhou, S.K.: Uod: Universal one-shot detection of anatomical landmarks. In: International Conference on Medical Image Computing and Computer-Assisted Intervention. pp. 24--34. Springer (2023)

\bibitem{zhu2021you}
Zhu, H., Yao, Q., Xiao, L., Zhou, S.K.: You only learn once: Universal anatomical landmark detection. In: Medical Image Computing and Computer Assisted Intervention--MICCAI 2021: 24th International Conference, Strasbourg, France, September 27--October 1, 2021, Proceedings, Part V 24. pp. 85--95. Springer (2021)

\end{thebibliography}

\end{document}